\PassOptionsToPackage{pdfpagelabels=false}{hyperref} 
\documentclass{IOS-Book-Article}
\usepackage{times}  
\usepackage{amsmath,amssymb}
\usepackage{booktabs}
\usepackage[T1]{fontenc}
\usepackage{bm}
\usepackage[toc]{appendix}
\usepackage{epsfig}

\DeclareMathOperator*{\argmax}{arg\,max}
\DeclareMathOperator*{\argmin}{arg\,min}
\DeclareMathOperator{\EX}{\mathbb{E}}

\usepackage{marginnote}
\usepackage[status=draft,lang=spanish, layout={marginclue, footnote}, mode=multiuser]{fixme}
\fxusetheme{color}
\FXRegisterAuthor{jb}{ajb}{Javier}
\FXRegisterAuthor{fn}{afn}{Fernando}
\FXRegisterAuthor{ja}{aja}{Jamie}

\normalfont

\begin{document}
\begin{frontmatter}       

\title{Synthetic ECG Generation for Data Augmentation and Transfer Learning in Arrhythmia Classification}

\author[A]{José Fernando Núñez}
\author[A]{Jamie Arjona\thanks{Corresponding Author: 
Jamie Arjona Martínez, Universitat Politècnica de Catalunya,  E-mail:
jamie.arjona@upc.edu.}}
\author[A]{Javier Béjar}

\address[A]{Computer Science Department, Universitat Politècnica de Catalunya (Barcelona Tech),  Barcelona, Spain}

\begin{abstract}
Deep learning models need a sufficient amount of data in order to be able to find the hidden patterns in it. It is the purpose of generative modeling to learn the data distribution, thus allowing us to sample more data and augment the original dataset. In the context of physiological data, and more specifically electrocardiogram (ECG) data, given its sensitive nature and expensive data collection, we can exploit the benefits of generative models in order to enlarge existing datasets and improve downstream tasks, in our case, classification of heart rhythm. 

In this work, we explore the usefulness of synthetic data generated with different generative models from Deep Learning namely Diffweave, Time-Diffusion and Time-VQVAE in order to obtain better classification results for two open source multivariate ECG datasets. Moreover, we also investigate the effects of transfer learning, by fine-tuning a synthetically pre-trained model and then progressively adding increasing proportions of real data. We conclude that although the synthetic samples resemble the real ones, the classification improvement when simply augmenting the real dataset is barely noticeable on individual datasets, but when both datasets are merged the results show an increase across all metrics for the classifiers when using synthetic samples as augmented data. From the fine-tuning results the Time-VQVAE generative model has shown to be superior to the others but not powerful enough to achieve results close to a classifier trained with real data only. In addition, methods and metrics for measuring closeness between synthetic data and the real one have been explored as a side effect of the main research questions of this study.
\end{abstract}

\end{frontmatter}

\section{Introduction}
The analysis of physiological data, often captured as time series (e.g., EEG, ECG, IMU, EMG), plays a crucial role in healthcare applications such as disease diagnosis and anomaly detection (\cite{wangSystematicReviewTime2022}). However, researchers face a significant challenge: the scarcity of publicly available datasets due to privacy concerns, ethical regulations, and the small number of patients with rare conditions (\cite{howeiiiEthicalChallengesPosed2020, federerBiomedicalDataSharing2015}). This low availability of data hinders the training of powerful deep learning models, which are known to be "data-hungry" (\cite{shaikhinaHandlingLimitedDatasets2017}). Furthermore, real-world physiological data often exhibit class imbalances in which the categories of interest are represented by far fewer examples compared to healthy conditions. This imbalance poses a challenge for machine learning algorithms, as they tend to favor the majority class during training (\cite{ishwaranCommentaryProblemClass2021}).

An approach to address both data scarcity and class imbalance is data augmentation. Traditional augmentation techniques, commonly used in computer vision (scaling, flipping, cropping) or adapted for time series (noise injection, jittering, warping), are often limited in their effectiveness (\cite{talavera2022data}). These methods often create data that deviates from the underlying distribution of the real data or simply represent minor modifications of the original samples.

Recently, generative modeling have become a very active area of research in the context of Deep Learning and offers a more promising approach for data augmentation across various domains, including text-to-image generation, natural language processing and time series generation (\cite{manduchiChallengesOpportunitiesGenerative2024}). However, applying generative models to physiological time series poses unique challenges as compared to images or text, since the usage of generative models for physiological time series (EEG, ECG, IMU, EMG, etc.) remains relatively unexplored.

This work presents an effort to address the aforementioned problems. We explore the selection of a suitable generative model for physiological time series data and propose methods to evaluate the quality of synthetic data. 
Our evaluation will consider the impact of synthetic data augmentation on existing physiological datasets, involving pre-training a classification model with synthetic data and fine-tuning it with real data to assess the impact of the generated data towards classification.

\section{Generative Networks}
From a probabilistic perspective, the goal of data modeling in machine learning is to find the conditional distribution $p_{\theta}(\bm{c}|\bm{x})$ of a vector $\bm{c}$ given the value of a vector $\bm{x}$ of input features, where the observed variable, $\bm{x}$, is a random sample of an unknown underlying process and $\theta$ the parameters of the chosen model. For classification tasks, $\bm{c}$ represents a discrete class label, and for regression, it corresponds to one or more continuous variables. The discriminative approach represents this conditional distribution with a parametric model and then finds the parameters using a training set of available data $({\bm{x}_{n},\bm{c}_{n}})$, which are pairs of input vectors with their associated target values. Given new values of $\bm{x}$, the learned conditional distribution can be used to make predictions of $\bm{c}$.

An alternative approach, albeit more difficult, is the generative one. The goal is to find the joint distribution $p_{\theta}(\bm{x},\bm{c})$, expressed, for example, as an implicit or explicit parametric model. Once learned, this distribution is used to evaluate the conditional probability $p_{\theta}(\bm{c}|\bm{x})$ in order to make predictions of $\bm{c}$ for new values of $\bm{x}$. This method assumes that it is possible to generate synthetic examples of the feature vector $\bm{x}$, given that the model learnt to capture correctly the process that defines the data. This approach is attractive, as it allows us to do sampling in order to obtain more data examples, i.e. new vectors $\bm{x}$, which is helpful when we have underrepresented classes in the training data or not enough data examples altogether.

There are many deep learning approaches to modeling data generatively as a probability distribution. Some families of models define proper probability distributions directly from the attributes of data, for example autoregressive models~\cite{bond2021deep} and normalizing flows~\cite{gui2020review}. Autoregressive models are a class of likelihood models that model the data distribution by estimating the data density. They approach the maximum likelihood objective $\theta^{*} = \argmax_{\theta} \EX_{x\sim p_{data}(\bm{x})}[\log p_{\theta}(\bm{x})]$ by factorizing $p_{\theta}(\bm{x})$ over the dimensions of $\bm{x}$ using the chain rule~\cite{dalal2019autoregressive}. Normalizing flows, on the other hand, provide a general way of constructing probability distributions over continuous random variables. The main idea is to express $\bm{x}$ as a transformation $T$ of a real vector $\bm{u}$ sampled from $p_{u}(\bm{u})$, with the condition that $T$ is a diffeomorphism, meaning that it is invertible and both $T$ and its inverse $T^{-1}$ are differentiable. Under these conditions, the density of $\bm{x}$ is well-defined and can be obtained by using a change of variables~\cite{papamakarios2021normalizing}.

Other families of models define the probability distributions by learning latent variables, such as VAEs~\cite{bond2021deep} and Probabilistic Denoising Diffusion~\cite{yang2022diffusion}. VAEs consist of an encoder that takes data $\bm{x}$ as input and transforms them into a latent code $\bm{z}$ of a space with less dimension than the input space, and a decoder, which takes a latent representation of the input data $\bm{z}$ and returns a reconstruction $\hat{\bm{x}}$. Variational inference is then performed to approximate the posterior of the model, $p_{\theta}(\bm{z}|\bm{x})$ so as to maximize the variational lower bound~\cite{kingma2022autoencoding}. Furthermore, Probabilistic Denoising Diffusion models assume that a network can be trained to learn to progressively remove noise from the data $\bm{x}$. This noise has been added applying a diffusion process that incrementally adds Gaussian noise using a variance schedule that destroys its information until it consists of unit Gaussian noise. They are of the form $p_{\theta}(\bm{x}_{0}) := \int p_{\theta}(\bm{x}_{0:T}) d \bm{x}_{1:T}$, where $\bm{x}_{1}, ... , \bm{x}_{T}$ are latents of the same dimensionality as the data $\bm{x}_{0} \sim q(\bm{x}_0)$. The joint distribution $p_{\theta}(\bm{x}_{0:T})$ is called the \textit{reverse process} and it is usually defined as a Markov chain with learned Gaussian transitions. The model is fitted using score matching~\cite{yang2022diffusion}.

Finally, models such as Generative Adversarial Networks (GANs) \cite{gui2020review} capture the data distribution implicitly and provide a way of interacting with the probability distribution indirectly. GANs learn by comparing new data samples with the original training data, learning to distinguish between data samples that are likely to be drawn from the same distribution as the training set and samples that are not. They do so by framing the data generation problem as a game between a network that can obtain samples from a latent Gaussian vector (generator) and a network whose task is to distinguish real from generated samples (discriminator). More specifically, to learn the generator's distribution $p_{g}$ over the data $\bm{x}$, a prior is defined on the input noise variables $p_{\bm{z}}(\bm{z})$, then a mapping to the data space is represented as $G(\bm{z};\theta_{g})$, where $G$ is a differentiable function with parameters $\theta_{g}$. A second network called the discriminator, represented by $D(\bm{x};\theta_{d})$, is defined to produce a number between 0 and 1, representing the probability that $\bm{x}$ came from training data rather than $p_{g}$. $D$ is trained to maximize the probability of assigning the correct label to both training examples and samples from $G$, while $G$ itself is simultaneously trained to minimize $\log (1-D(G(\bm{z})))$~\cite{goodfellow2014generative}. 

\section{ECG related tasks literature review}
\subsection{ECG generation}
The recording of heart electrical activity is represented as time series data, that is, realizations of random variables indexed by time. These types of data can be found in a wide range of domains: physiological processes ~\cite{andrzejak_indications_2001}, stock markets ~\cite{zhao_stock_2023}, industry applications ~\cite{kashpruk_time_2023}, climate change ~\cite{mudelsee_trend_2019}, ITS services ~\cite{arjona_martinez_2021}, etc. each one having its own problems and attributes. Thus, the literature for time series modeling has been a very active research area for more than fifty years. Recently, with the rise of Deep Learning methods, there has been an increasing interest in the application of generative models in some of these domains, especially those in which there is a small amount of data, or small representation of the signals of interest.

Within the literature of deep learning, time series generation approaches can be categorized principally by two key aspects: data utilization strategies and employed model architectures. Data utilization strategies encompass univariate, multivariate, non-conditional, and conditional approaches. Conversely, model architectures encompass diverse architectures such as Generative Adversarial Networks (GANs), Variational Autoencoders (VAEs) and Denoising Diffusion models.

 For ECG data, data are usually registered using 12 sensors called leads, that capture a different direction of cardiac activation in 3D space. For univariate data, studies employ a single signal, the most commonly used is lead II. These studies may work directly with the full signal or with windows corresponding to individual heartbeats. GAN approaches in the literature use Convolutional Neural Network (CNN), Recurrent Neural Networks (RNN) or a combination of both for the generator and the discriminator architectures. As such we find the work of ~\cite{zhu_electrocardiogram_2019} that proposes a Bi-Long Short Term Memory architecture (Bi-LSTM) to unconditionally learn from single leads. Furthermore, \cite{brophy_quick_2019} proposes a Wasserstein GAN with gradient penalty which transforms unconditioned signal data into images to use as input to a 2D CNN generator and then translates the images back to time series data, although information is lost in these transformations. Moreover, ~\cite{delaney_synthesis_2019} proposes GAN models which  explore two different kinds of generators (LSTM and Bi-LSTM) and also explores two different types of discriminators, LSTM and CNN, with the goal of unconditionally synthesizing the lead II signal by segmenting it into individual heart beats. In ~\cite{golany_pgans_2019}, the authors propose using a CNN GAN model for each patient to eliminate the need to use conditioned data and then train an LSTM classifier with both real and synthetic data. The data used consisted of heartbeats from a single channel. The authors of ~\cite{wang_ecg_2019} proposed to use a 1D-CNN Conditional GAN in order to gconditionally generate lead II data with the class label, and then use this data to improve the classification of an LSTM model trained with the generated and real data simultaneously. Similarly, ~\cite{nankani_investigating_2020} investigates conditioning the generator of a deep convolutional GAN to achieve the generation of three distinct heartbeat conditions using a segmented heartbeat approach. However, their work also highlights limitations associated with this method, such as the convergence challenges faced when training conditioned GAN models. The work of~\cite{xia_ecg_2023} proposes and compares two new approaches to the generation of single heartbeat signal from lead II conditioned on four different classes by using a conditional VAE and a conditional Wasserstein GAN and claims superior performance from the VAE. Motivated by the promising performance of probabilistic diffusion models, the authors of~\cite{adib_synthetic_2023} propose the application of an improved Denoising Diffusion Probabilistic Model (DDPM) for ECG data generation. They compare their approach to a Wasserstein GAN with gradient penalty, employing a technique called Gramian Angular Summation/Difference Fields and Markov Transition Fields to embed the 1D ECG signals into a 2D space. Notably, this work focuses on generating signals from the normal class using a single channel, and the authors report that the GAN model achieved superior performance in this specific case.

Multivariate works make use of multiple leads and focus on using the entire signal, as opposed to segmenting it into individual heartbeats. The authors of the work ~\cite{zhang_synthesis_2021} use a 2D bi-LSTM GAN in order to produce synthetic standard 12-lead ECGs data conditioned on four different types of heart anomalies, claiming high quality on the generated data and being the first to use a large sample of data coming from different datasets: PTB-XL ~\cite{Wagner2020-PTBXL}, CCDD ~\cite{zhang_CCDD_2010}, CSE, Chapman ~\cite{hangyuan_CHAPMAN_2019} and also a private dataset. Building upon this research direction, ~\cite{thambawita_deepfake_2021} has become a highly influential work for benchmarking state of the art deep learning methods in ECG data generation. The authors propose to unconditionally generate 10 second signals from 12-lead normal condition ECGs using WaveGAN*, a model originally intended for audio data, and also introduced a new model called Pulse2Pulse which is a modification of WaveGAN*. Authors from ~\cite{brophy_multivariate_2021} also propose the use of multiple channels of data in order to improve the generation of synthetic ECGs, using an unconditional GAN made up of an LSTM generator and a CNN discriminator with a novel Dynamic Time Warping loss. Alternative approaches, such as the work by ~\cite{li_tts-cgan_2022} have explored the application of Transformers, proposing a novel GAN architecture called TTS-CGAN, which incorporates Transformers in both the generator and discriminator. This architecture enables the generation of conditioned synthetic ECG signals segmented into individual heartbeats. DDPM models have also been used for multivariate ECG signal generation mainly combined with new architectures based on State Space Models (SSM). The works in this line of research are ~\cite{alcaraz_diffusion-based_2023} which propose the Structural State Space diffusion ECG (SSSD-ECG) model that uses data from the PTB-XL dataset to generate conditioned synthetic ECGs and compares their results with those obtained by the models WaveGAN* and Pulse2Pulse from ~\cite{thambawita_deepfake_2021}. While the proposed diffusion-based model demonstrates superior performance compared to the benchmarks, the authors acknowledge that the synthetic samples still exhibit limitations in fully replicating the quality of real data. This finding suggests a gap between the characteristics of real and synthetic ECG data, highlighting a research area which needs further improvement. Following this, ~\cite{zama_ecg_2023} also proposes a DDPM with a state space augmented transformer architecture. Using the PTB-XL dataset, their model uses 12-lead ECG signals to generate synthetic data of 12 different classes. The authors make use of a combination of SSM layers in order to capture global patterns and components from a transformer to capture local patterns, and while achieving superior results compared to the benchmarks (SSSD-ECG, WaveGAN*, and Pulse2Pulse), but the authors acknowledge that the generated data remains insufficient to train a classifier which rivals one trained with real ECG data.

\begin{table}
\centering
\caption{\label{tab:deep_learning_models} Comparison of Deep Learning Models for ECG Generation}
\begin{tabular}{llrrr}
\toprule
 \textbf{Data type}  &   \textbf{Model type} &  \textbf{References}  \\
\midrule
\textbf{univariate (unconditioned)} & GAN      & \cite{zhu_electrocardiogram_2019,brophy_quick_2019,delaney_synthesis_2019,golany_pgans_2019}    \\
             & DDDPM &  \cite{adib_synthetic_2023} \\
\textbf{univariate (conditioned)} & GAN      &  \cite{wang_ecg_2019,nankani_investigating_2020,xia_ecg_2023} \\
             & VAE &  \cite{xia_ecg_2023} \\
\textbf{multivariate (unconditioned)} & GAN      &  \cite{thambawita_deepfake_2021, brophy_multivariate_2021} \\
\textbf{multivariate (conditioned)} & GAN      &  \cite{zhang_synthesis_2021, li_tts-cgan_2022} \\
             & DDPM &  \cite{alcaraz_diffusion-based_2023, zama_ecg_2023} \\
\bottomrule
\end{tabular}
\end{table}

\subsection{ECG classification}
Evaluation of heart conditions is the main purpose of the ECG recordings and there is much interest in automating this process through modeling, especially for the detection of arrhythmia conditions. The release of the MIT-BIH dataset marked a significant turning point, allowing researchers to explore automated ECG analysis using statistical, machine learning and data mining techniques. The proposal in \cite{dechazalAutomaticClassificationHeartbeats2004} is one of the pioneering works in introducing a statistical model (linear discriminant) for ECG classification that achieved significant precision and recall in the problem. In the same year, \cite{osowskiSupportVectorMachinebased2004} introduced Support Vector Machines and demonstrated a significant improvement in accuracy when compared to previous works, making SVM the state of the art learning algorithm for this task. Early research also explored other modeling techniques like decision trees \cite{luzECGArrhythmiaClassification2013}, KNN clustering \cite{luzECGArrhythmiaClassification2013} and logistic regression \cite{escalona-moranElectrocardiogramClassificationUsing2015}. These advancements showcased the potential of ML for automating ECG analysis.

Beyond statistical and machine learning models, there is another branch of research that leverages deep learning models such as CNNs and recurrent neural networks (RNNs) for ECG tasks. The works of \cite{zhangPatientSpecificECGClassification2017,faustAutomatedDetectionAtrial2018,luiMulticlassClassificationMyocardial2018,sanninoDeepLearningApproach2018,xiangAutomaticQRSComplex2018,chenAutomatedArrhythmiaClassification2020} or \cite{mahmudDeepArrNetEfficientDeep2020} all proposed different types of DL architectures that obtained the highest accuracy score on the MIT-BIH arrhythmia dataset for multi-class classification. These approaches often involve preprocessing the ECG into PQRST complexes prior to classification.

Recently, the availability of new ECG datasets with expanded features has fueled further advances in deep learning for ECG analysis. These datasets offer increased numbers of patients, leads, and classes, leading to higher data dimensionality (both \textit{temporal} and \textit{spatial}). The increased spatial dimension arises from the inclusion of more leads, which captures parallel electrical activity within more regions of the heart. \cite{strodthoff_ptbxl_benchmark_2020} reviewed classification methods for ECG on the PTB-XL and ICBEB2018 datasets and categorized them in the following groups: CNN architectures (full convolution and Resnets with 2D CNNs), RNN architectures (LSTM and GRU and variants like biLSTM and biGRU), and baseline classifiers (Wavelet + shallow NN or naive forecasts).

\section{Datasets}
A variety of ECG datasets can be found in the literature, most of which are for public use. These are characterized mainly by the sample frequency, the length of the signal, and the number of leads. Some of the first datasets were focused in providing samples from specific heart diseases from a small set of patients and using a small number of leads like in \cite{moody_impact_2001} but recent ones follow the format offered in the Chinese Cardiovascular Disease Database ~\cite{zhang_CCDD_2010} in which there are a wide variety of signals of heart conditions and heart diseases using 12 leads, 500hz or 100hz sample frequencies and signal duration of 10 seconds, like for example in the CHAPMAN dataset ~\cite{hangyuan_CHAPMAN_2019}.
Due to the need of large number of examples that offer a great and rich variety of samples for the different ECG classes, current literature on ECG generation and classification uses the most recent datasets available on the literature, i.e. the PTB-XL~\cite{Wagner2020-PTBXL}, the CHAPMAN~\cite{hangyuan_CHAPMAN_2019}, the CCDD~\cite{zhang_CCDD_2010} and the ICBEB~\cite{liu_ICBEB_2018}. For a more exhaustive review of the subject, we refer the reader to the following articles \cite{xiao_classification_review_2023}, \cite{ansari_classification_review_2023} and \cite{strodthoff_ptbxl_benchmark_2020}.

In this work, we use the PTB-XL and the CHAPMAN dataset as these are not only the most recent ones, but they also offer a rich variety of samples related to heart rhythms for a large number of different classes. A summary of the datasets used is shown in \ref{datasets_table}
\subsection{PTB-XL: Physikalisch-Technische Bundesanstalt XL}
The PTB-XL electrocardiogram (ECG) dataset, introduced by \cite{Wagner2020-PTBXL}, has been created and organized towards the use of Deep Learning and Machine Learning models to push forward the state of the art on ECG-related tasks. This dataset includes ECG recordings from a large population of 18,885 patients. Each recording utilizes the standard 12-lead configuration and possesses an average duration of approximately 10 seconds. Notably, the data are available in two sampling frequencies: 100 Hz and 500 Hz, catering to different research needs. The PTB-XL dataset offers a rich annotation scheme, employing a hierarchical classification system with 71 distinct classes, facilitating the classification of ECG signals at various levels of granularity. The hierarchy categorizes the classes into superclasses and subclasses. Superclasses represent broader categories of ECG findings, including normal, conduction disturbance, myocardial infarction, hypertrophy, and ST/T wave changes. Each superclass is further subdivided into more specific subclasses that pinpoint the precise underlying pathology. Furthermore, the dataset also includes a metadata file that contains detailed information on patients' demographics and diagnoses. This information enhances the utility of the dataset for studies investigating the relationship between ECG findings and clinical conditions. Furthermore, PTB-XL provides a classification scheme based on the physiological origin of the ECG abnormality, categorizing classes by form (e.g. QRS complex morphology) or rhythm (e.g. atrial fibrillation).
\subsection{12-lead electrocardiogram - Chapman University and Shaoxing People’s Hospital}
The Chapman electrocardiogram (ECG) dataset, introduced by ~\cite{zheng_12-lead_CHAPMAN_2020} offers another valuable resource for the analysis of cardiac rhythms. This dataset comprises ECG recordings from a substantial patient population of 45,152 patients. Each recording utilizes the standard 12-lead configuration and possesses an average duration of approximately 10 seconds with a sampling frequency of 500 Hz, which is in line with the typical parameters used in the acquisition of ECG. The Chapman dataset specifically focuses on the classification of heart rhythm abnormalities. It defines eleven distinct classes encompassing various rhythm conditions, including sinus bradycardia, sinus rhythm, atrial fibrillation, sinus tachycardia, atrial flutter, sinus irregularity, supra-ventricular tachycardia, atrioventricular nodal re-entrant tachycardia, atrioventricular re-entrant tachycardia, and sinus arrhythmia with atrial wander. Notably, a significant overlap exists between these arrhythmia classes and those offered by the PTB-XL dataset. This compatibility facilitates the potential for combining these resources in order to obtain a more rich dataset as some of the classes have more representation in one of the datasets compared to the other.

\begin{table}
\centering
\caption{\label{datasets_table} Summary of the datasets used.}
\begin{tabular}{llrrr}
\toprule
\textbf{Dataset}& \textbf{\# Patients}&  \textbf{\# Leads} &  \textbf{Average length (seconds)}&  \textbf{\# Classes} \\
\midrule
\textbf{PTB-XL} & 18885      &  12  &  10   & 71 \\
\textbf{CHAPMAN}& 45152      &  12  &  10   & 11 \\
\bottomrule
\end{tabular}
\end{table}

\section{Models}
\subsection{Generative models}

For the purpose of trying out different approaches to generative modeling, we have chosen to train three state of the art models, with the condition that the model's implementation was open sourced, and that we were able to train the models successfully without major modifications or extensive hyperparameter tuning. Here we provide a short description of each model.

Denoising Diffusion models, given their immense popularity, it is no surprise these are also considered state of the art in synthetic time series generation. We opted for three different architectures, namely Diffwave, \cite{kong_diffwave_2021} Diffusion-TS \cite{yuan2024diffusion} and the Unet1D conditional model from Huggingface. 
Diffwave, originally intended for audio data, is a non-autoregressive diffusion probabilistic model which supports conditional waveform generation. The model learns to convert white noise into a structured signal using a Markov chain with a constant number of steps at synthesis. It uses a variation of the variational lower bound as an optimization function on the data log likelihood, and supports \textit{local conditioning} on linguistic features, the mel spectrogram, or the hidden states of the text to wave architecture. Furthermore, there is a \textit{global conditioner} given by a discrete label, in which a shared embedding is used with dimension $ d_{\text{label}}  = 128$ in the original experiments \cite{kong_diffwave_2021}. The SSSD-ECG proposed in ~\cite{alcaraz_diffusion-based_2023} is used in this work instead of the original Diffwave.

The Unet1D model from Huggingface's Diffusers library is based on the U-Net model, introduced in \cite{ronneberger2015unet}. The U-Net architecture was created to address the challenge of limited annotated data in the medical field. It is composed of a contracting and an expansive path; the contracting path has encoder layers that capture contextual information and reduce the dimension of the input, while the expansive path works with decoder layers which learn to expand the dimension of the encoded data via skip connections to generate a segmentation map \cite{ronneberger2015unet}. Given that the original implementation of the Unet1D diffusion model was not conditioned, we needed to condition it ourselves following the same conditioning scheme as in the Unet2D conditioned model, also provided by the Diffusers library. Please note that in the rest of this paper we refer to this model as \textit{Time-Diffusion}.


Time-VQVAE, presented in \cite{lee2023vector}, is a two-step modeling approach similar to the one in \cite{oord2017vqvae} in which a VQ-VAE is used for the first stage and MaskGIT for the second. VQ-VAE acts on the frequency domain, separating the signals into their low and high frequency components after applying the short-term Fourier Transform (STFT). Then, two sets of encoder, decoder, and vector-quantizer are used to learn the discrete latent spaces for LF and HF, and afterwards, priors of these spaces are learned with two bidirectional Transformer models. Lastly, the model jointly and conditionally samples sets of LF and HF discrete latent vectors from the learned priors and decodes them into the time-domain with the learned decoders. 

\subsection{Classification model}

\begin{figure}[b]
\begin{center}
    \includegraphics[angle=90, trim={2cm 0cm  0cm 0cm},clip, width=\textwidth]{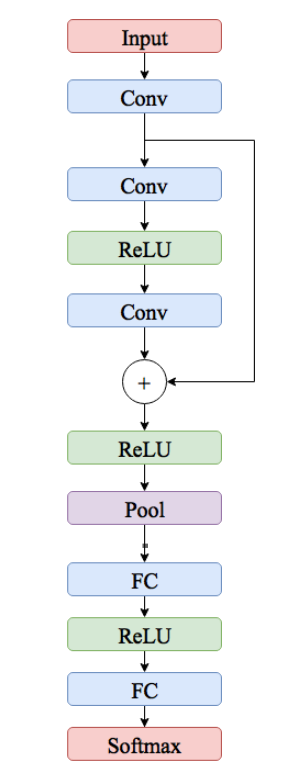}
\end{center}
\caption{\label{fig:classifier} 1 dimensional deep residual convolutional classifier architecture.}
\end{figure}

In order to evaluate the quality of the generated data in our experiments and its effect on classification tasks, this work proposes to use a 1D CNN model similar to a ResNet classifier (\cite{wang_time_2016}). This architecture has been extensively used in the ECG classification literature, \cite{strodthoff_ptbxl_benchmark_2020} and has become the standard model for this kind of problems. The input of the model has shape $(bs, c, h, w)$ where $bs$ is the batch size, $c$ is the number of channels, $h$ the height of the signal and $w$ the number of steps in the signal. We have considered $h=1$ and $c$ to be the number of signals from the ECG so the input becomes a tensor of shape $(bs, c, w)$. The first layer of the model consists in a 1d convolutional layer that keeps the length of the input signal. Then, a set of convolutional blocks consisting on 1d convolutional layers with residual connections and with ReLU activations functions are considered. Following the convolutional blocks, a set of dense layers are used in order to provide the final output of the model. The values for the hyperparameters have been obtained by using a tree-structured parzen estimator algorithm (\cite{optuna_2019}). Hyperparameters are different based on the dataset used as shown in \autoref{tab:classifier_hparams}. A simplified diagram of the model is shown in \autoref{fig:classifier}.

\begin{table}
\centering
\caption{\label{tab:classifier_hparams} Obtained hyperparameters values of the 1d cnn model for each dataset.}
\begin{tabular}{llr}
\toprule
\textbf{Dataset}& \textbf{Hyperparameter}   &   \textbf{Value} \\
\midrule
\textbf{PTB-XL} &                      &               \\
                & n. conv blocks       & 6             \\
                & n. kernels           & 32            \\
                & filter size          & 7x7           \\
                & n. neurons classifier& 256           \\
                & n. layers classifier & 3             \\
                & learning rate        & 0.000354      \\
                & dropout rate         & 0.474333      \\
                & earlystop patience   & 10            \\
                & earlystop delta      & 0.00001       \\
\textbf{CHAPMAN}&                      &               \\
                & n. conv blocks       & 5             \\
                & n. kernels           & 16            \\
                & filter size          & 7x7           \\
                & n. neurons classifier& 256           \\
                & n. layers classifier & 3             \\
                & learning rate        & 0.000158      \\
                & dropout rate         & 0.377601      \\
                & earlystop patience   & 10            \\
                & earlystop delta      & 0.00001       \\
\bottomrule
\end{tabular}
\end{table}

\section{Methodology}
This work focuses on the generation of realistic electrocardiogram (ECG) signals. To achieve this goal, we leverage two established ECG datasets: PTB-XL and Chapman. Each dataset offers unique characteristics and information about cardiac electrical activity. To fully exploit this data diversity, we employ a tailored approach.

For each dataset, a distinct generative model will be trained. This isolates the model's learning process on the specific features and statistical properties present in each dataset. Subsequently, a comprehensive evaluation will be conducted to assess the quality and realism of the synthetic ECG representations generated for each individual dataset.

Following these individual evaluations, we will explore the potential benefits of combining the information from both the PTB-XL and Chapman datasets. A new, unified dataset will be constructed by merging the ECG signals from both sources. This enriched dataset will encompass a broader range of ECG signal variations and potentially offer a more comprehensive representation of human cardiac electrical activity. Leveraging this newly created dataset, we will train new generative models. These models will be tasked with learning the underlying patterns and statistical relationships within the combined data, aiming to generate even more realistic and informative synthetic ECG signals.

\section{Evaluation}\label{sec:evaluation}
Evaluating the quality of generated time series data remains an open challenge, unlike synthetic images where established metrics like Fréchet Inception Distance (FID) \cite{bynagariGANsTrainedTwo2019} or Maximum Mean Discrepancy (MMD) \cite{liMMDGANDeeper} exists. The concept of "natural" time series and pre-trained feature extraction networks, crucial for FID in images, isn't directly applicable here. Additionally, capturing the underlying information in real data through the generated data distribution presents difficulties, especially for conditional generation tasks. Due to the lack of a way to assess the quality of the generated data, we explore different heuristics with different hypothesis:
\begin{enumerate}
    \item Visualizations: Use of dimensionality reduction techniques like t-Distributed Stochastic Nieghbor Embedding (t-SNE) \cite{JMLR:v9:vandermaaten08a}, Uniform Manifold Approximation and Projection (UMAP) \cite{McInnes2018} and Pairwise Controlled Manifold Approximation projection (PaCMAP) \cite{JMLR:wang_pacmap_2021}. The goal is to validate through visualizations in a two dimensional space if the distribution of the generated synthetic data and the real data cluster together.
    \item Metrics:
    \begin{itemize}
        \item 2-Sample test classification score: A classifier trained with real data is used in order to discriminate between real data and synthetic data. If the classifier cannot discern real from synthetic data then the accuracy score will be 50\%. If the classifier is able to perfectly discriminate real from synthetic, its accuracy will be close to 100\%.
        \item MMD is a non-parametric two sample test to determine if samples are drawn from the same distribution. The Radial Basis Function kernel is used in order to compute the distance between two different projections of the mean in a very high (it could be infinite) dimensional space. The lower the value, the closer the distributions of the real and the synthetic data.        
    \end{itemize}
    \item Evaluation metrics from classifier trained under different metrics as explained in \ref{classifier-eval}.
\end{enumerate}

\subsection{Classifier evaluations}\label{classifier-eval}
As an evaluation approach, we have chosen to analyze changes in classification performance metrics (accuracy, precision, recall and roc-auc) through training of a classifier (or a pre-trained classifier on synthetic data for the transfer learning task) on different combinations of real and synthetic data for the train and the evaluation: 
\begin{enumerate}
    \item \textbf{Training-Testing Split variations}:
    \begin{itemize}
        \item \textbf{TrRTeR}: Train and evaluate on real data.
        \item \textbf{TrSTeS}: Train and evaluate on synthetic data.
        \item \textbf{TrSTeR}: Train on synthetic data and evaluate on real data.
        \item \textbf{TrRTeS}: Train on real data and evaluate on synthetic data.
        \item \textbf{TrRSTeR}: Train on a mix of real and synthetic data and evaluate on real data.
    \end{itemize}
    \item \textbf{Transferability}: Train on synthetic. Then perform fine-tuning of the model by freezing all the layers except the ones of the classifier (dense layers near the output of the model) and add different proportions of real data. The goal is to evaluate the amount of real samples needed to successfully tune the synthetically pre-trained model and achieve an equal or better performance metric than the model trained only with real data.
\end{enumerate}

These evaluations aim to understand how different training procedures on real and synthetic data impact classification tasks. The ultimate goal is to improve the classifier's performance with synthetic data as opposed to using only real data during training.

\section{Experiments}
In order to evaluate the synthetic data, the same experiments are performed on three ECG datasets: \textit{PTB-XL}, \textit{CHAPMAN} and a merge of \textit{PTB-XL} with \textit{CHAPMAN}.

The first experimental stage consists on obtaining a classifier that is able to discern the patterns of different ECG classes considered in the real data and find the hyperparameters of the model that will be used with different combinations of the real data and synthetic data. This means that a subset of classes is selected in order to make proper comparison between the PTB-XL and CHAPMAN datasets. The selected classes are sinus bradycardia (SBRAD), sinus rhythm (SR), atrial fibrilation (AFIB), sinus tachycardia (STACH), atrial flutter (AFLT), sinus arrhythmia (SARRH) and supraventricular tachycardia (SVTAC). Table \ref{tab:datasets_classes_representation} shows the representation (number of samples) for each class. The goal of this stage is to obtain the best values of the hyperparameters of the model for the real data in order to fix it for the rest of models that will be trained in stage 2. 

\begin{table}
\centering
\caption{\label{tab:datasets_classes_representation} Classes and their representation in each dataset.}
\begin{tabular}{llll}
\toprule
\textbf{Class}& \textbf{ID} & \textbf{PTB-XL}   & \textbf{CHAPMAN}\\
\midrule
SBRAD   & 0& 509    & 12792 \\
SR      & 1& 13404  & 6306 \\
AFIB    & 2& 1210   & 1441 \\
STACH   & 3& 643    & 5726 \\
AFLT    & 4& 34     & 6218 \\
SARRH   & 5& 613    & 2119 \\
SVTAC   & 6& 20     & 501 \\
\end{tabular}
\end{table}

Then, as explained in section \ref{sec:evaluation}, new classifiers are trained and tested with different combinations of real and synthetic data in order to assess the quality of the synthetic data obtained by different generative models. The performance of the models on the \textbf{TrSTeS}, \textbf{TrSTeR} and \textbf{TrRTeS} combinations will indicate how closely related are the synthetic and real data. From the \textbf{TrSTeS} setting it is expected that the classifier obtains a similar score in all metrics to a classifier trained and evaluated on the real data only. From the \textbf{TrSTeR} and \textbf{TrRTeS} experiments it is expected to have similar scores in all metrics. Finally, the \textbf{TrRSTeR} setting can be used as an indicative on how the synthetic data influences the final model, helping it in order to obtain better classification scores or worsening it. Aside to the evaluation of synthetic data from each generative model, an additional synthetic dataset is created by merging together samples from the generative models in order to balance possible differences between those.

Finally, fine-tuning experiments are performed in order to assess how much real data is needed in a model trained with synthetic data only in order to obtain the same results as a model trained with real data. The chunks of real data considered are of size 20\%, 40\%, 60\% and 80\%. Note that a 100\% real data added would not be equivalent to the \textbf{TrRSTeR} experiment setting as in the fine-tuning experiments the model is not being re-trained, as only the last layers (those tagged as classifier layers) are the ones that are being re-trained with a small learning rate.

In all the experiments the metrics considered are the (accuracy/ precision/recall/f1-score/ROC AUC) average values from repeating the same experiment $n=25$ times in order to reduce variability between executions. The ROC AUC metric has been added in order to get an approximated comparison with other benchmarks on the used datasets (e.g.: \cite{strodthoff_ptbxl_benchmark_2020}).

\section{Results}

\subsection{PTB-XL experiments}
Table \ref{tab:test_table_PTB-XL} shows the results of the experiments performed on the \textit{PTB-XL} dataset. The first row shows the results of a classifier trained and evaluated with real data only. 

In general, classifiers trained on synthetic data have good scores when evaluated on synthetic data but in the hybrid settings (\textbf{TrSTeR} and \textbf{TrRTeS}) classifiers get a very low score across all the metrics. For the \textbf{TrRSTeR} case the metrics are close to those obtained with models trained only with real data.

If we analyze the results for each different generative model we observe that classifiers using \textit{diffwave} synthetic data are the ones that obtain all scores closer to the real data in the \textbf{TrRSTeR} but are also the ones with the lower scores in the other settings. The classifiers that used the \textit{Time-Diffusion} synthetic data achieve a precision score ($0.6823$) better than the real data classifier ($0.6485$) in the \textbf{TrRTeS} setting, this means that the number of false positives is lower when compared with a classifier trained with the real data only and could be an indicative that the generated synthetic data is very similar to the real data but with lower noise, thus the classification of the synthetic data is easier than classify real data. In the case of the classifiers trained with synthetic data from the \textit{Time-VQVAE} model, we observe that in the \textbf{TrSTeS} setting the trained classifiers had a harder time to classify the samples, presenting lower scores than those classifiers trained on the synthetic data of the other generative models. In the \textbf{TrSTeR} setting the classifiers outperformed all the others in the recall metric ($0.6793$), this means that the number of false negatives has been reduced and the reason could be that the generated data is similar to the real data in each of the classes and caused a positive effect increasing the representation of the low represented classes. Finally, for the classifiers trained with a merge of all the synthetic generated samples, the scores obtained reflect clearly a leverage between the differences between each one of the individual generative models as these are neither better nor worse than any other of the individual models.

Table \ref{tab:fine-tune_PTB-XL_v2} shows the results of the fine-tuning experiments performed on the \textit{PTB-XL} dataset. A model trained on synthetic data only is fine tuned with different percentages of real data from the PTB-XL dataset. Scores from a model trained only with real data with the same percentage is shown in order to evaluate the impact of the synthetic data. 
It is clear that the classifier pre-trained with the \textit{Time-VQVAE} synthetic data outperforms all the other classifiers and this suggests that the quality of the \textit{Time-VQVAE} samples is higher than those samples from the other methods. In all the cases except for the \textbf{Excution Time} it is observed that the models fall short behind the scores of the model trained with real data only. Finally, observe that similar to what has been observed in \ref{tab:test_table_PTB-XL}, the model trained with a merge of samples from all the generative model just leverages the scores from the individual cases.

\begin{table}
\centering
\caption{\label{tab:test_table_PTB-XL} Results from testing the models under different train and test settings using the PTB-XL dataset.}
\begin{tabular}{lllrrrrr}
\toprule
\textbf{Dataset}&\textbf{Generative model}&\textbf{Setting}&\textbf{Accuracy} &\textbf{Precision}&\textbf{Recall}&\textbf{f1-score}&\textbf{ROC AUC}\\
\midrule
\textbf{PTB-XL} & -          &\textbf{TrRTeR}    & 0.9208  & 0.6485   & 0.6202  & 0.6194 & 0.9530   \\
    & \textit{diffwave}      & \textbf{TrSTeS}   & 0.9367  & 0.9378   & 0.9362  & 0.9358 & 0.9927   \\
    & \textit{diffwave}      & \textbf{TrSTeR}   & 0.3492  & 0.238    & 0.2386  & 0.1830 & 0.7166   \\
    & \textit{diffwave}      & \textbf{TrRTeS}   & 0.3657  & 0.473    & 0.3598  & 0.3071 & 0.7939   \\
    & \textit{diffwave}      & \textbf{TrRSTeR}  & 0.9208  & 0.6369   & 0.5946  & 0.5975 & 0.9549  \\
    & \textit{Time-Diffusion}& \textbf{TrSTeS}   & 0.9673  & 0.9676   & 0.9669  & 0.9666 & 0.9962   \\
    & \textit{Time-Diffusion}& \textbf{TrSTeR}   & 0.6383  & 0.3468   & 0.553   & 0.3758 & 0.8608   \\
    & \textit{Time-Diffusion}& \textbf{TrRTeS}   & 0.5652  & 0.6823   & 0.5695  & 0.5142 & 0.9396   \\
    & \textit{Time-Diffusion}& \textbf{TrRSTeR}  & 0.9088  & 0.6114   & 0.575   & 0.5674 & 0.9489   \\
    & \textit{Time-VQVAE}    & \textbf{TrSTeS}   & 0.7163  & 0.7178   & 0.714   & 0.7122 & 0.9358   \\
    & \textit{Time-VQVAE}    & \textbf{TrSTeR}   & 0.7333  & 0.3916   & 0.6793  & 0.4265 & 0.8905   \\
    & \textit{Time-VQVAE}    & \textbf{TrRTeS}   & 0.4024  & 0.4512   & 0.4018  & 0.3538 & 0.8267   \\
    & \textit{Time-VQVAE}    & \textbf{TrRSTeR}  & 0.9104  & 0.6213   & 0.6007  & 0.594  & 0.9490   \\
    & \textit{all}     & \textbf{TrSTeS}   & 0.86    & 0.8614 & 0.8591 & 0.8587 & 0.9794  \\
    & \textit{all}     & \textbf{TrSTeR}   & 0.6738  & 0.3815 & 0.6796 & 0.4207 & 0.8946   \\
    & \textit{all}     & \textbf{TrRTeS}   & 0.4216  & 0.6119 & 0.4205 & 0.3781 & 0.8542  \\
    & \textit{all}     & \textbf{TrRSTeR}  & 0.904   & 0.6021 & 0.6243 & 0.5884 & 0.9403   \\ 
\bottomrule
\end{tabular}
\end{table}

\begin{table}
\centering
\caption{\label{tab:fine-tune_PTB-XL_v2} Results of the transferability capabilities between a model trained on synthetic data and fine-tuned on real data from the PTB-XL dataset.  Different percentages of real data are used in order to perform the fine-tuning. Highlighted in bold the best score across the metrics when comparing individual generative methods}
\begin{tabular}{llrrrrr}
\toprule
\textbf{Metric}& \textbf{Data \%}  & \textbf{Real} &  \textbf{Diffwave} &  \textbf{Time-diffusion}&  \textbf{Time-VQVAE} &\textbf{all} \\
\midrule
\textbf{Accuracy}   
            & 0.2 &0.8743  &0.7416   &0.3471  &\textbf{0.7946}  &0.5747  \\
            & 0.4 &0.8957  &0.787    &0.2844  &\textbf{0.8358}  &0.6823  \\
            & 0.6 &0.9091  &0.8204   &0.3247  &\textbf{0.8353}  &0.7305  \\
            & 0.8 &0.9186  &0.8252   &0.4401  &\textbf{0.8566}  &0.7466  \\
            & 1   &0.9208  &0.8266   &0.4726  &\textbf{0.8622}  &0.7924  \\
\textbf{Precision}  
            & 0.2 &0.5485  &0.3161   &0.2928  &\textbf{0.3946}  &0.3339  \\
            & 0.4 &0.5984  &0.434    &0.3756  &\textbf{0.463 } & 0.3557 \\
            & 0.6 &0.6066  &0.4753   &0.3196  &\textbf{0.4287}  &0.3533  \\
            & 0.8 &0.6467  &0.4612   &0.3563  &\textbf{0.4573}  &0.3626  \\
            & 1   &0.6485  &0.4802   &0.3915  &\textbf{0.4854}  &0.3837  \\
\textbf{Recall}     
            & 0.2 &0.6521  &0.3413   &0.4038  &\textbf{0.4082}  &0.4183  \\
            & 0.4 &0.6184  &0.3082   &0.4045  &\textbf{0.4155}  &0.4145  \\
            & 0.6 &0.6322  &0.2908   &\textbf{0.4239}  &0.4158  &0.4191  \\
            & 0.8 &0.632   &0.2895   &\textbf{0.4051}  &0.3849  &0.4124  \\
            & 1   &0.6202  &0.2721   &\textbf{0.4115}  &0.3977  &0.4017  \\
\textbf{f1-score}   
            & 0.2 &0.5745  &0.2939   &0.2368  &\textbf{0.3786}  &0.2938  \\
            & 0.4 &0.5896  &0.3322   &0.2692  &\textbf{0.4281}  &0.3145  \\
            & 0.6 &0.6008  &0.3327   &0.241   &\textbf{0.4104}  &0.3278  \\
            & 0.8 &0.6204  &0.3227   &0.2904  &\textbf{0.4076}  &0.3424  \\
            & 1   &0.6194  &0.3076   &0.3173  &\textbf{0.4258}  &0.3581  \\
\textbf{ROC AUC}    
            & 0.2 &0.9549  &0.7227   &0.831   &\textbf{0.9104}  &0.8256  \\
            & 0.4 &0.9566  &0.7197   &0.8195  &\textbf{0.9119}  &0.8036  \\
            & 0.6 &0.9619  &0.7739   &0.8012  &\textbf{0.9088}  &0.8356  \\
            & 0.8 &0.9548  &0.7812   &0.8075  &\textbf{0.9047}  &0.8333  \\
            & 1   &0.953   &0.7753   &0.8048  &\textbf{0.9058}  &0.8415  \\
                    
\textbf{Execution Time (sec)}   
            & 0.2 &9.75    &8.4433   &6.0378   &\textbf{3.6742 }   &4.6459 \\
            & 0.4 &14.7862 &10.6608  &7.1733   &\textbf{7.6464 }   &7.8385 \\
            & 0.6 &21.7929 &13.3107  &9.0513   &\textbf{9.9524 }   &10.044 \\
            & 0.8 &27.6365 &16.3846  &14.0942  &\textbf{12.4406}   &15.594  \\
            & 1   &37.2186 &21.3287  &21.2917  &\textbf{18.6273}   &22.0278  \\
\bottomrule
\end{tabular}
\end{table}

\subsection{CHAPMAN experiments} 
Following the same structure from the previous subsection, Table \ref{tab:test_table_CHAPMAN} shows the results of the experiments performed on the \textit{CHAPMAN} dataset where the first row corresponds to the model trained with real data only.

In general, the scores in the \textbf{TrRSTeR} setting from each generative model are very close to the scores of the real data only model.

For each specific generative model, we observer that \textit{diffwave} falls behind in all the scores in the hybrid settings and scores better in the synthetic only setting, \textbf{TrSTeS}. In the case of \textit{Time-Diffusion} the scores are behind those obtained with the real data only model but higher than the classifiers trained on \textit{diffwave} samples, observe that in the \textbf{TrRTeS} setting the scores obtained by the trained classifiers are higher than any other of the ones trained on data from the other generative models indicating a similarity between the real and the synthetic data. From the classifiers that used synthetic data from \textit{Time-VQVAE} we observe that, similar to the PTB-XL results, they scored the lowest in the \textbf{TrSTeS}. From the \textbf{TrSTeR} setting it can be observed that the \textit{Time-VQVAE} samples have a high score for the recall metric (lower number of false positives) suggesting that the increased number of samples for the low represented classes has a positive effect on the classifiers. This can also be in line with the low scores reached on the \textbf{TrRTeS} setting: the increased number of samples for the low represented classes in the evaluation have a negative effect on the classifiers trained on the real data only. Finally, for the classifiers that use a merge of the synthetic samples from all the generative models, we observe that scores are leveraged between the low values and the high values of the scores of the individual cases.

Table \ref{tab:fine-tune_CHAPMAN_v2} shows the results of the fine-tuning experiments performed on the \textit{CHAPMAN} dataset. A model trained on synthetic data only is fine tuned with different percentages of real data. Scores from a model trained only with real data with the same percentage is shown for comparisons. 

Similar to the results presented in Table \ref{tab:fine-tune_PTB-XL_v2} we observe that the \textit{Time-VQVAE} model obtains the highest scores across all metrics when compared to the other generative models but values are lower when compared to the real data only model. The only metric where generative models excels is in the execution: starting with a pre-trained model on synthetic data helps to converge faster and also that the resources needed are lower than when training with real data only.

\begin{table}
\centering
\caption{\label{tab:test_table_CHAPMAN} Results from testing the models under different train and test combinations with synthetic data from different sources.}
\begin{tabular}{lllrrrrr}
\toprule
\textbf{Dataset}& \textbf{Generative model}  & \textbf{Setting} &  \textbf{Accuracy} &  \textbf{Precision}&  \textbf{Recall} & \textbf{f1-score} & \textbf{ROC AUC}\\
\midrule
    \textbf{CHAPMAN}& -     & \textbf{TrRTeR}   &0.8688  &0.6876  &0.734   &0.7072  &0.9615  \\
    & \textit{diffwave}     & \textbf{TrSTeS}   &0.8931  &0.8944  &0.8924  &0.8921  &0.9803  \\
    & \textit{diffwave}     & \textbf{TrSTeR}   &0.4208	 &0.435   &0.3947  &0.3205  &0.7882  \\
    & \textit{diffwave}     & \textbf{TrRTeS}   &0.4327	 &0.544	  &0.4414  &0.3814  &0.8355  \\
    & \textit{diffwave}     & \textbf{TrRSTeR}  &0.8672	 &0.6876  &0.723   &0.7018  &0.9602  \\
    & \textit{Time-Diffusion}& \textbf{TrSTeS}  &0.9057  &0.908  &0.9067  &0.9062  &0.9776  \\
    & \textit{Time-Diffusion}& \textbf{TrSTeR}  &0.6668  &0.5513  &0.5677  &0.5338  &0.8958  \\
    & \textit{Time-Diffusion}& \textbf{TrRTeS}  &0.5835  &0.5813  &0.5929  &0.5481  &0.9112  \\
    & \textit{Time-Diffusion}& \textbf{TrRSTeR} &0.8587  &0.6993  &0.7248  &0.7046  &0.9584  \\
    & \textit{Time-VQVAE}    & \textbf{TrSTeS}  &0.7131  &0.6973  &0.7028  &0.692   &0.918  \\
    & \textit{Time-VQVAE}    & \textbf{TrSTeR}  &0.7053  &0.5678  &0.6514  &0.5592  &0.9162  \\
    & \textit{Time-VQVAE}    & \textbf{TrRTeS}  &0.4463  &0.5172  &0.4426  &0.3731  &0.8594  \\
    & \textit{Time-VQVAE}    & \textbf{TrRSTeR} &0.8589  &0.6952  &0.7188  &0.6962  &0.9571  \\
    & \textit{all}           & \textbf{TrSTeS}  &0.822	 &0.8208  &0.8204  &0.8191  &0.9634 \\
    & \textit{all}           & \textbf{TrSTeR}  &0.7346	 &0.6059  &0.6861  &0.5903  &0.9307 \\
    & \textit{all}           & \textbf{TrRTeS}  &0.4878	 &0.5666  &0.4925  &0.4436  &0.8705 \\
    & \textit{all}           & \textbf{TrRSTeR} &0.8514  &0.682   &0.7142  &0.6874  &0.9552 \\     
\bottomrule
\end{tabular}
\end{table}

\begin{table}
\centering
\caption{\label{tab:fine-tune_CHAPMAN_v2} Study of the transferability capabilities between a model trained on synthetic data and fine-tuned on real data vs model trained with real data. Different percentages of real data are used. The data used corresponds to the CHAPMAN dataset and the synthetic data used comes from all the generative models. Highlighted in bold the best score across the metric when comparing individual generative methods}
\begin{tabular}{llrrrrr}
\toprule
\textbf{Metric}& \textbf{Data \%}  & \textbf{Real} &  \textbf{Diffwave} &  \textbf{Time-diffusion}&  \textbf{Time-VQVAE} &\textbf{all} \\
\midrule
\textbf{Accuracy}   
            & 0.2 &0.8344  &0.6093   &0.6618   &\textbf{0.7214} &0.8018    \\
            & 0.4 &0.8536  &0.6334   &0.6613   &\textbf{0.7222} &0.7876    \\
            & 0.6 &0.8627  &0.6317   &0.67     &\textbf{0.7207} &0.8091    \\
            & 0.8 &0.867   &0.639    &0.6708   &\textbf{0.7227} &0.8072    \\
            & 1   &0.8688  &0.6428   &0.6786   &\textbf{0.7263} &0.8078    \\
\textbf{Precision}  
            & 0.2 &0.6466  &0.4983   &0.5091   &\textbf{0.586 } &0.6394    \\
            & 0.4 &0.6683  &0.4921   &0.5218   &\textbf{0.5673} &0.6309    \\
            & 0.6 &0.6809  &0.4528   &0.536    &\textbf{0.5658} &0.6621    \\
            & 0.8 &0.6836  &0.4548   &0.5254   &\textbf{0.5633} &0.6559    \\
            & 1   &0.6876  &0.4773   &0.5264   &\textbf{0.5644} &0.6602    \\
\textbf{Recall}     
            & 0.2 &0.6893  &0.3775   &0.4446   &\textbf{0.4966} &0.6013    \\
            & 0.4 &0.7172  &0.4082   &0.4585   &\textbf{0.5115} &0.5904    \\
            & 0.6 &0.7229  &0.4052   &0.4668   &\textbf{0.5122} &0.6064    \\
            & 0.8 &0.7278  &0.4248   &0.4646   &\textbf{0.5117} &0.6053    \\
            & 1   &0.734   &0.4176   &0.4658   &\textbf{0.5178} &0.6118    \\
\textbf{f1-score}  
            & 0.2 &0.6571  &0.3772   &0.4376   &\textbf{0.4816} &0.6022    \\
            & 0.4 &0.6889  &0.4034   &0.4484   &\textbf{0.4971} &0.5866    \\
            & 0.6 &0.697   &0.4029   &0.4549   &\textbf{0.5002} &0.6167    \\
            & 0.8 &0.7028  &0.417    &0.4561   &\textbf{0.499 } &0.6108    \\
            & 1   &0.7072  &0.4147   &0.4633   &\textbf{0.5057} &0.6204    \\
\textbf{ROC AUC}    
            & 0.2 &0.9334  &0.8286   &0.8473   &\textbf{0.8989} &0.9301    \\
            & 0.4 &0.9481  &0.8426   &0.8624   &\textbf{0.9107} &0.9297    \\
            & 0.6 &0.952   &0.8314   &0.8687   &\textbf{0.8994} &0.9351    \\
            & 0.8 &0.9591  &0.8438   &0.8687   &\textbf{0.9053} &0.9325    \\
            & 1   &0.9615  &0.8418   &0.872    &\textbf{0.9083} &0.9353    \\
                    
\textbf{Execution Time (sec)}   
            & 0.2 &23.8719  &12.5884   &13.635  &\textbf{10.5368} &10.596      \\
            & 0.4 &42.2111  &25.139    &22.993  &\textbf{15.3488} &14.9487    \\
            & 0.6 &60.6021  &28.7951   &30.4444 &\textbf{24.709 } &27.234    \\
            & 0.8 &74.7595  &36.2134   &34.795  &\textbf{27.8938} &32.3562    \\
            & 1   &107.3153 &54.7783   &49.5382 &\textbf{43.6309} &44.6556    \\
\bottomrule
\end{tabular}
\end{table}

\subsection{PTB-XL + CHAPMAN experiments}
Finally, the experiments are repeated on a merge of both datasets, PTB-XL and CHAPMAN. From this new dataset the low represented classes from each dataset will increase their representation because PTB-XL and CHAPMAN have different number of samples in each class~(\ref{tab:datasets_classes_representation} and it is expected to have an effect on the distributions learn by the generative models.
The results from these experiments are presented in Table~\ref{tab:test_table_BOTH}.

In general, classifiers trained on synthetic data perform slightly better than the real data only classifiers across all metrics under the \textbf{TrRSTeR} setting. The possible reason of this change when compared to the other datasets could be the increased representation of the low represented classes. This result is also observed in the classifiers trained with the merge of synthetic samples from the different generative models.

Inspecting the models one by one we can see similar patterns as in the previous datasets. The \textit{diffwave} model is the one with the lower scores across the hybrid settings and has high scores on the \textbf{TrSTeS} setting. It can be observed that \textit{Time-Diffusion} and \textit{Time-VQVAE} present similar scores but the latter has a higher score for the recall metric ($0.5983$) in the \textbf{TrSTeR} setting. This value is far from the real data only model ($0.7526$) but is way higher than the other generative models ($0.4612$ \textit{Diffweave} and $0.5053$ \textit{Time-Diffusion}). This increase on recall (meaning a lower false positives) is consistent with the previous datasets for the \textit{Time-VQVAE} model and the \textbf{TrSTeR} setting and clearly indicates that the \textit{Time-VQVAE} samples have higher representation quality.

Table~\ref{tab:fine-tune_BOTH_v2} shows the results of the fine-tuning experiments performed on the \textit{PTB-XL+CHAPMAN} dataset. As before, a model trained on synthetic data only is fine tuned with different percentages of real data. Scores from a model trained only with real data with the same percentages is shown for comparison. Similar than in the previous cases, it can be seen that the \textit{Time-VQVAE} excels the others in nearly all the metrics except in accuracy, where \textit{Time-Diffusion} presents better results. Although the classifiers achieve better scores with the \textit{Time-VQVAE} compared when the data comes from the other generative models, the results are far away from the scores obtained with real data only classifiers except for the Execution Time, where the difference in training time is more or less half the time in the \textit{Time-VQVAE} when compared with the real data classifiers.

\begin{table}
\centering
\caption{\label{tab:test_table_BOTH} Results from testing the models under different train and test combinations.}
\begin{tabular}{lllrrrrr}
\toprule
\textbf{Dataset}& \textbf{Generative model}  & \textbf{Setting} &  \textbf{Accuracy} &  \textbf{Precision}&  \textbf{Recall} & \textbf{f1-score} & \textbf{ROC AUC}\\
\midrule
    \textbf{PTBXL+CHAPMAN}& -     & \textbf{TrRTeR} &0.8574 &0.751  &0.7526 &0.7475 &0.9683 \\
    & \textit{diffwave}     & \textbf{TrSTeS}       &0.9395 &0.9407 &0.9396 &0.9394 &0.9928 \\
    & \textit{diffwave}     & \textbf{TrSTeR}       &0.6297 &0.4859 &0.4612 &0.4492 &0.8289 \\
    & \textit{diffwave}     & \textbf{TrRTeS}       &0.4693 &0.5853 &0.4708 &0.4411 &0.8815 \\
    & \textit{diffwave}     & \textbf{TrRSTeR}      &0.8673 &0.77   &0.7573 &0.7613 &0.9741 \\
    & \textit{Time-Diffusion}& \textbf{TrSTeS}      &0.946  &0.9473 &0.946  &0.946  &0.9939 \\
    & \textit{Time-Diffusion}& \textbf{TrSTeR}      &0.4353 &0.5177 &0.5053 &0.4243 &0.8423 \\
    & \textit{Time-Diffusion}& \textbf{TrRTeS}      &0.6647 &0.7203 &0.6577 &0.6553 &0.9395 \\
    & \textit{Time-Diffusion}& \textbf{TrRSTeR}     &0.865  &0.76   &0.759  &0.7567 &0.9735 \\
    & \textit{Time-VQVAE}    & \textbf{TrSTeS}      &0.6353 &0.6313 &0.6383 &0.6303 &0.8759 \\
    & \textit{Time-VQVAE}    & \textbf{TrSTeR}      &0.621  &0.5123 &0.5983 &0.4927 &0.8756 \\
    & \textit{Time-VQVAE}    & \textbf{TrRTeS}      &0.335  &0.4587 &0.3303 &0.3163 &0.7562 \\
    & \textit{Time-VQVAE}    & \textbf{TrRSTeR}     &0.8607 &0.7453 &0.778  &0.7567 &0.9733 \\
    & \textit{all}           & \textbf{TrSTeS}      &0.8271 &0.8292 &0.8281 &0.8268 &0.9702 \\
    & \textit{all}           & \textbf{TrSTeR}      &0.6741 &0.5598 &0.6445 &0.5552 &0.9133 \\
    & \textit{all}           & \textbf{TrRTeS}      &0.4955 &0.5795 &0.493 &0.4888 &0.8683 \\
    & \textit{all}           & \textbf{TrRSTeR}     &0.8614 &0.754 &0.7619 &0.7543 &0.9738 \\     
\bottomrule
\end{tabular}
\end{table}

\begin{table}
\centering
\caption{\label{tab:fine-tune_BOTH_v2} Study of the transferability capabilities between model trained on synthetic data and fine-tuned on real data vs model trained with real data  using different percentages of real data. The data used corresponds to the PTB-XL+CHAPMAN dataset. Highlighted in bold the best score across the metric when comparing individual generative methods}
\begin{tabular}{llrrrrr}
\toprule
\textbf{Metric}& \textbf{Data \%}  & \textbf{Real} &  \textbf{Diffwave} &  \textbf{Time-Diffusion}&  \textbf{Time-VQVAE}& \textbf{all} \\
\midrule
\textbf{Accuracy}   
            & 0.2 &0.7973  &0.5636   &0.6261          &\textbf{0.6433} &0.7081    \\
            & 0.4 &0.8162  &0.5947   &\textbf{0.6632} &0.622	       &0.7188    \\
            & 0.6 &0.8272  &0.5955   &\textbf{0.673}  &0.6339	       &0.7326    \\
            & 0.8 &0.8439  &0.5633   &\textbf{0.6734} &0.6448	       &0.7345    \\
            & 1   &0.8574  &0.6048   &\textbf{0.677}  &0.6464	       &0.7405    \\
\textbf{Precision}  
            & 0.2 &0.6767  &0.4749  &0.5159  &\textbf{0.5775}    &0.5706    \\
            & 0.4 &0.7062  &0.4778  &0.5215  &\textbf{0.5731}    &0.5647    \\
            & 0.6 &0.7172  &0.49    &0.524   &\textbf{0.5865}    &0.5772    \\
            & 0.8 &0.7383  &0.4968  &0.5236  &\textbf{0.5811}    &0.5894    \\
            & 1   &0.751   &0.5032  &0.5269  &\textbf{0.5838}    &0.5916   \\
\textbf{Recall}     
            & 0.2 &0.6853  &0.4076   &0.457	  &\textbf{0.5299}   &0.5804	    \\
            & 0.4 &0.7186  &0.4366   &0.485	  &\textbf{0.5239}   &0.5727	    \\
            & 0.6 &0.7327  &0.4523   &0.4908  &\textbf{0.5252}   &0.5803	    \\
            & 0.8 &0.7448  &0.4434   &0.4942  &\textbf{0.5282}   &0.5802	    \\
            & 1   &0.7526  &0.4526   &0.4911  &\textbf{0.5209}   &0.5792	    \\
\textbf{f1-score}   
            & 0.2 &0.6752  &0.3742   &0.4578  &\textbf{0.5082} &0.5703   \\
            & 0.4 &0.7062  &0.4118   &0.4877  &\textbf{0.5077} &0.5571    \\
            & 0.6 &0.7205  &0.4274   &0.4912  &\textbf{0.5043} &0.5706    \\
            & 0.8 &0.7375  &0.4107   &0.493	  &\textbf{0.5136} &0.5727    \\
            & 1   &0.7475  &0.4348   &0.4912  &\textbf{0.513 } &0.5724    \\
\textbf{ROC AUC}    
            & 0.2 &0.9333  &0.8121   &0.8561  &\textbf{0.8702} &0.884    \\
            & 0.4 &0.9507  &0.8162   &0.8623  &\textbf{0.879 } &0.8775    \\
            & 0.6 &0.9581  &0.8296   &0.8683  &\textbf{0.8748} &0.8861    \\
            & 0.8 &0.9662  &0.8266   &0.8641  &\textbf{0.88  } &0.8865    \\
            & 1   &0.9683  &0.8396   &0.8683  &\textbf{0.8815} &0.8873    \\
                    
\textbf{Execution Time (sec)}
            & 0.2 &31.7487	 &19.437  &23.1133  &\textbf{16.9806}	&17.3352  \\
            & 0.4 &59.9915	 &41.2056 &39.2475  &\textbf{31.0467}	&33.6572  \\
            & 0.6 &82.0948	 &55.2982 &59.6008  &\textbf{38.5391}	&45.2609  \\
            & 0.8 &119.2049  &62.3021 &69.802   &\textbf{46.8767}	&56.778	  \\
            & 1   &151.2782  &81.879  &87.2528  &\textbf{62.2831}	&68.0647  \\
\bottomrule
\end{tabular}
\end{table}

\section{Conclusions}
\subsection{The effect of synthetic data. It helps?}
The results obtained in the experiments highlight that generated samples with current state of the art methods from Deep Learning cannot be used as standalone data in order to substitute real data. When considering the use of synthetic data as a technique to increase the number of samples of low represented classes, the conclusions are similar when considering only the \textbf{PTB-XL} or \textbf{CHAPMAN} datasets individually, but results from the experiments on the \textbf{PTB-XL+CHAPMAN} datasets show an increase in all metrics under the \textbf{TrRSTeS} setting, pointing towards a better quality of the synthetic data that helps to improve (slightly) the classifiers' performance.

For the transfer learning experiments it can be said that there is not a clear advantage on using a pre-trained model on synthetic data as classifiers being fine-tuned with all the real data are not able to get scores as high as the model trained with all the real data. This is observed across all the studied datasets and settings.

\subsection{About the generative models. Is there a clear winner?}
Although the quality of the generated samples does not match the real data, experiments show that \textit{diffwave} samples seem worse than those from \textit{Time-Diffusion} and \textit{Time-VQVAE}. For \textit{Time-VQVAE} and \textit{Time-Diffusion} is somehow difficult to evaluate if one is superior than the other in the experiments with different settings of real and synthetic data, but the fine-tuning experiments highlight a superiority in the samples from \textit{Time-VQVAE} in all the scores by a large amount. Also it has been observed that the \textit{Time-VQVAE} samples help to reduce the number of false positives in the \textit{TrSTeR} setting across all the datasets.

\subsection{Methods to differentiate real from synthetic. Which one to use?}
Different techniques to differentiate real from synthetic samples suggested in the literature have been tested: MMD, 2-sample test, reduction-based visualization techniques and classifiers. From those it has been observed that MMD and visualization techniques lack the power to differentiate synthetic data from real one as the results suggest that synthetic and real data share the same distribution. On the other hand, evaluation methods using classifiers, 2-sample test and classifiers under different settings, have shown that there is indeed a difference between synthetic and real data. The problem of using classifiers to assess the quality of the data is the lack of a function and the need to fit multiple models, something that could be not possible in some problems.

\subsection{Next steps}
This work highlights different problems in the current literature in order to work with synthetic data that tries to resemble periodic Time Series like ECG but in particular there is one that is more important than any other: the lack of a metric to be able to quantify the difference between the real and the synthetic data. Such quantification will be of much help in detecting authorship from popular generative models like ChatGPT as highlighted in \cite{liang2024syntheticdatareviews}.

It has been observed that the merge of the PTB-XL and CHAPMAN datasets has helped to improve the classifiers when trained in the \textbf{TrRSTeR} setting, and interpreting such setting as applying data augmentation. This highlights that using a bigger dataset could help in order to create an ECG generative model able to generate samples with good enough quality so that they can be used for data augmentation in other tasks.

Finally, it would be of interest to replicate the experiments presented in this work with other types of physiological data such as EEGs and compare the results about the quality of the representations. By doing so, we can compare the performance of other synthetically augmented datasets, and thus assess whether the current results are explained by the nature of the multivariate ECG data, or if they also occur in other types of physiological data. 


\section{Acknowledgments}

 This research has been funded by the Artificial Intelligence for Healthy Aging (AI4HA, MIA.2021.M02.0007) project from the Programa Misiones de I+D en Inteligencia Artificial 2021 and by the European Union-NextGenerationEU, Ministry of Universities and Recovery, Transformation and Resilience Plan, through a call from Universitat Politècnica de Catalunya (Grant Ref. 2021UPC-MS-67461).

\bibliographystyle{ieeetr}
\bibliography{paper}

\begin{thebibliography}{10}

\bibitem{wangSystematicReviewTime2022}
W.~K. Wang, I.~Chen, L.~Hershkovich, J.~Yang, A.~Shetty, G.~Singh, Y.~Jiang, A.~Kotla, J.~Z. Shang, R.~Yerrabelli, A.~R. Roghanizad, M.~M.~H. Shandhi, and J.~Dunn, ``A {{Systematic Review}} of {{Time Series Classification Techniques Used}} in {{Biomedical Applications}},'' {\em Sensors}, vol.~22, p.~8016, Jan. 2022.

\bibitem{howeiiiEthicalChallengesPosed2020}
E.~G. Howe~III and F.~Elenberg, ``Ethical {{Challenges Posed}} by {{Big Data}},'' {\em Innov Clin Neurosci}, vol.~17, pp.~24--30, Oct. 2020.

\bibitem{federerBiomedicalDataSharing2015}
L.~M. Federer, Y.-L. Lu, D.~J. Joubert, J.~Welsh, and B.~Brandys, ``Biomedical {{Data Sharing}} and {{Reuse}}: {{Attitudes}} and {{Practices}} of {{Clinical}} and {{Scientific Research Staff}},'' {\em PLOS ONE}, vol.~10, p.~e0129506, June 2015.

\bibitem{shaikhinaHandlingLimitedDatasets2017}
T.~Shaikhina and N.~A. Khovanova, ``Handling limited datasets with neural networks in medical applications: {{A}} small-data approach,'' {\em Artificial Intelligence in Medicine}, vol.~75, pp.~51--63, Jan. 2017.

\bibitem{ishwaranCommentaryProblemClass2021}
H.~Ishwaran and R.~O'Brien, ``Commentary: {{The Problem}} of {{Class Imbalance}} in {{Biomedical Data}},'' {\em J Thorac Cardiovasc Surg}, vol.~161, pp.~1940--1941, June 2021.

\bibitem{talavera2022data}
E.~Talavera, G.~Iglesias, {\'A}.~Gonz{\'a}lez-Prieto, A.~Mozo, and S.~G{\'o}mez-Canaval, ``Data augmentation techniques in time series domain: A survey and taxonomy,'' {\em arXiv preprint arXiv:2206.13508}, 2022.

\bibitem{manduchiChallengesOpportunitiesGenerative2024}
L.~Manduchi, K.~Pandey, R.~Bamler, R.~Cotterell, S.~D{\"a}ubener, S.~Fellenz, A.~Fischer, T.~G{\"a}rtner, M.~Kirchler, M.~Kloft, Y.~Li, C.~Lippert, G.~{de Melo}, E.~Nalisnick, B.~Ommer, R.~Ranganath, M.~Rudolph, K.~Ullrich, G.~V. den Broeck, J.~E. Vogt, Y.~Wang, F.~Wenzel, F.~Wood, S.~Mandt, and V.~Fortuin, ``On the {{Challenges}} and {{Opportunities}} in {{Generative AI}},'' Feb. 2024.

\bibitem{bond2021deep}
S.~Bond-Taylor, A.~Leach, Y.~Long, and C.~G. Willcocks, ``Deep generative modelling: A comparative review of vaes, gans, normalizing flows, energy-based and autoregressive models,'' {\em IEEE transactions on pattern analysis and machine intelligence}, 2021.

\bibitem{gui2020review}
J.~Gui, Z.~Sun, Y.~Wen, D.~Tao, and J.~Ye, ``A review on generative adversarial networks: Algorithms, theory, and applications,'' {\em IEEE Transactions on Knowledge and Data Engineering}, vol.~35, no.~4, pp.~3313--3332, 2023.

\bibitem{dalal2019autoregressive}
M.~Dalal, A.~C. Li, and R.~Taori, ``Autoregressive models: What are they good for?,'' 2019.

\bibitem{papamakarios2021normalizing}
G.~Papamakarios, E.~Nalisnick, D.~J. Rezende, S.~Mohamed, and B.~Lakshminarayanan, ``Normalizing flows for probabilistic modeling and inference,'' {\em J. Mach. Learn. Res.}, vol.~22, jan 2021.

\bibitem{yang2022diffusion}
L.~Yang, Z.~Zhang, and S.~Hong, ``Diffusion models: A comprehensive survey of methods and applications,'' {\em arXiv preprint arXiv:2209.00796}, 2022.

\bibitem{kingma2022autoencoding}
D.~P. Kingma and M.~Welling, ``Auto-encoding variational bayes,'' 2022.

\bibitem{goodfellow2014generative}
I.~J. Goodfellow, J.~Pouget-Abadie, M.~Mirza, B.~Xu, D.~Warde-Farley, S.~Ozair, A.~Courville, and Y.~Bengio, ``Generative adversarial networks,'' 2014.

\bibitem{andrzejak_indications_2001}
R.~G. Andrzejak, K.~Lehnertz, F.~Mormann, C.~Rieke, P.~David, and C.~E. Elger, ``Indications of nonlinear deterministic and finite-dimensional structures in time series of brain electrical activity: dependence on recording region and brain state,'' {\em Physical Review. E, Statistical, Nonlinear, and Soft Matter Physics}, vol.~64, p.~061907, Dec. 2001.

\bibitem{zhao_stock_2023}
C.~Zhao, P.~Hu, X.~Liu, X.~Lan, and H.~Zhang, ``Stock {Market} {Analysis} {Using} {Time} {Series} {Relational} {Models} for {Stock} {Price} {Prediction},'' {\em Mathematics}, vol.~11, p.~1130, Jan. 2023.
\newblock Number: 5 Publisher: Multidisciplinary Digital Publishing Institute.

\bibitem{kashpruk_time_2023}
N.~Kashpruk, C.~Piskor-Ignatowicz, and J.~Baranowski, ``Time {Series} {Prediction} in {Industry} 4.0: {A} {Comprehensive} {Review} and {Prospects} for {Future} {Advancements},'' {\em Applied Sciences}, vol.~13, p.~12374, Jan. 2023.
\newblock Number: 22 Publisher: Multidisciplinary Digital Publishing Institute.

\bibitem{mudelsee_trend_2019}
M.~Mudelsee, ``Trend analysis of climate time series: {A} review of methods,'' {\em Earth-Science Reviews}, vol.~190, pp.~310--322, Mar. 2019.

\bibitem{arjona_martinez_2021}
J.~Arjona~Martinez, M.~P. Linares, and J.~Casanovas, ``Characterizing parking systems from sensor data through a data-driven approach,'' {\em Transportation Letters}, vol.~13, pp.~183--192, Mar. 2021.
\newblock Publisher: Taylor \& Francis \_eprint: https://doi.org/10.1080/19427867.2020.1866331.

\bibitem{zhu_electrocardiogram_2019}
F.~Zhu, F.~Ye, Y.~Fu, Q.~Liu, and B.~Shen, ``Electrocardiogram generation with a bidirectional {LSTM}-{CNN} generative adversarial network,'' {\em Sci Rep}, vol.~9, p.~6734, May 2019.
\newblock Number: 1 Publisher: Nature Publishing Group.

\bibitem{brophy_quick_2019}
E.~Brophy, Z.~Wang, and T.~E. Ward, ``Quick and {Easy} {Time} {Series} {Generation} with {Established} {Image}-based {GANs},'' Oct. 2019.
\newblock arXiv:1902.05624 [cs, stat].

\bibitem{delaney_synthesis_2019}
A.~M. Delaney, E.~Brophy, and T.~E. Ward, ``Synthesis of {Realistic} {ECG} using {Generative} {Adversarial} {Networks},'' Sept. 2019.
\newblock arXiv:1909.09150 [cs, eess, stat].

\bibitem{golany_pgans_2019}
T.~Golany and K.~Radinsky, ``Pgans: Personalized generative adversarial networks for ecg synthesis to improve patient-specific deep ecg classification,'' {\em Proceedings of the AAAI Conference on Artificial Intelligence}, vol.~33, pp.~557--564, July 2019.
\newblock Number: 01.

\bibitem{wang_ecg_2019}
P.~Wang, B.~Hou, S.~Shao, and R.~Yan, ``{ECG} {Arrhythmias} {Detection} {Using} {Auxiliary} {Classifier} {Generative} {Adversarial} {Network} and {Residual} {Network},'' {\em IEEE Access}, vol.~7, pp.~100910--100922, 2019.
\newblock Conference Name: IEEE Access.

\bibitem{nankani_investigating_2020}
D.~Nankani and R.~D. Baruah, ``Investigating {Deep} {Convolution} {Conditional} {GANs} for {Electrocardiogram} {Generation},'' in {\em 2020 {International} {Joint} {Conference} on {Neural} {Networks} ({IJCNN})}, pp.~1--8, July 2020.
\newblock ISSN: 2161-4407.

\bibitem{xia_ecg_2023}
Y.~Xia, W.~Wang, and K.~Wang, ``{ECG} signal generation based on conditional generative models,'' {\em Biomedical Signal Processing and Control}, vol.~82, p.~104587, Apr. 2023.

\bibitem{adib_synthetic_2023}
E.~Adib, A.~S. Fernandez, F.~Afghah, and J.~J. Prevost, ``Synthetic {ECG} {Signal} {Generation} {Using} {Probabilistic} {Diffusion} {Models},'' {\em IEEE Access}, vol.~11, pp.~75818--75828, 2023.

\bibitem{zhang_synthesis_2021}
Y.-H. Zhang and S.~Babaeizadeh, ``Synthesis of standard 12‑lead electrocardiograms using two-dimensional generative adversarial networks,'' {\em J Electrocardiol}, vol.~69, pp.~6--14, 2021.

\bibitem{Wagner2020-PTBXL}
P.~Wagner, N.~Strodthoff, R.-D. Bousseljot, W.~Samek, and T.~Schaeffter, ``{PTB-XL}, a large publicly available electrocardiography dataset,'' Apr. 2020.

\bibitem{zhang_CCDD_2010}
J.-w. Zhang, L.-p. Wang, X.~Liu, H.-h. Zhu, and J.~Dong, ``Chinese {Cardiovascular} {Disease} {Database} ({CCDD}) and {Its} {Management} {Tool},'' in {\em 2010 {IEEE} {International} {Conference} on {BioInformatics} and {BioEngineering}}, (Philadelphia, PA, USA), pp.~66--72, IEEE, May 2010.

\bibitem{hangyuan_CHAPMAN_2019}
G.~Hangyuan, ``A 12-lead electrocardiogram database for arrhythmia research covering more than 10,000 patients,'' Nov. 2019.
\newblock Publisher: figshare.

\bibitem{thambawita_deepfake_2021}
V.~Thambawita, J.~L. Isaksen, S.~A. Hicks, J.~Ghouse, G.~Ahlberg, A.~Linneberg, N.~Grarup, C.~Ellervik, M.~S. Olesen, T.~Hansen, C.~Graff, N.-H. Holstein-Rathlou, I.~Strümke, H.~L. Hammer, M.~M. Maleckar, P.~Halvorsen, M.~A. Riegler, and J.~K. Kanters, ``Deepfake electrocardiograms using generative adversarial networks are the beginning of the end for privacy issues in medicine,'' {\em Sci Rep}, vol.~11, p.~21896, Nov. 2021.
\newblock Number: 1 Publisher: Nature Publishing Group.

\bibitem{brophy_multivariate_2021}
E.~Brophy, M.~De~Vos, G.~Boylan, and T.~Ward, ``Multivariate {Generative} {Adversarial} {Networks} and {Their} {Loss} {Functions} for {Synthesis} of {Multichannel} {ECGs},'' {\em IEEE Access}, vol.~9, pp.~158936--158945, 2021.
\newblock Conference Name: IEEE Access.

\bibitem{li_tts-cgan_2022}
X.~Li, A.~H.~H. Ngu, and V.~Metsis, ``{TTS}-{CGAN}: {A} {Transformer} {Time}-{Series} {Conditional} {GAN} for {Biosignal} {Data} {Augmentation},'' June 2022.
\newblock arXiv:2206.13676 [cs].

\bibitem{alcaraz_diffusion-based_2023}
J.~M.~L. Alcaraz and N.~Strodthoff, ``Diffusion-based conditional {ECG} generation with structured state space models,'' {\em Computers in Biology and Medicine}, vol.~163, p.~107115, Sept. 2023.

\bibitem{zama_ecg_2023}
M.~H. Zama and F.~Schwenker, ``{ECG} {Synthesis} via {Diffusion}-{Based} {State} {Space} {Augmented} {Transformer},'' {\em Sensors}, vol.~23, p.~8328, Jan. 2023.
\newblock Number: 19 Publisher: Multidisciplinary Digital Publishing Institute.

\bibitem{dechazalAutomaticClassificationHeartbeats2004}
P.~{de Chazal}, M.~O'Dwyer, and R.~B. Reilly, ``Automatic classification of heartbeats using {{ECG}} morphology and heartbeat interval features,'' {\em IEEE Trans Biomed Eng}, vol.~51, pp.~1196--1206, July 2004.

\bibitem{osowskiSupportVectorMachinebased2004}
S.~Osowski, L.~T. Hoai, and T.~Markiewicz, ``Support vector machine-based expert system for reliable heartbeat recognition,'' {\em IEEE Trans Biomed Eng}, vol.~51, pp.~582--589, Apr. 2004.

\bibitem{luzECGArrhythmiaClassification2013}
E.~J. d.~S. Luz, T.~M. Nunes, V.~H.~C. {de Albuquerque}, J.~P. Papa, and D.~Menotti, ``{{ECG}} arrhythmia classification based on optimum-path forest,'' {\em Expert Systems with Applications}, vol.~40, pp.~3561--3573, July 2013.

\bibitem{escalona-moranElectrocardiogramClassificationUsing2015}
M.~A. {Escalona-Moran}, M.~C. Soriano, I.~Fischer, and C.~R. Mirasso, ``Electrocardiogram {{Classification Using Reservoir Computing With Logistic Regression}},'' {\em IEEE J. Biomed. Health Inform.}, vol.~19, pp.~892--898, May 2015.

\bibitem{zhangPatientSpecificECGClassification2017}
C.~Zhang, G.~Wang, J.~Zhao, P.~Gao, J.~Lin, and H.~Yang, ``Patient-{{Specific ECG Classification Based}} on {{Recurrent Neural Networks}} and {{Clustering Technique}},'' in {\em Biomedical {{Engineering}}}, ACTA Press, Mar. 2017.

\bibitem{faustAutomatedDetectionAtrial2018}
O.~Faust, A.~Shenfield, M.~Kareem, T.~R. San, H.~Fujita, and U.~R. Acharya, ``Automated detection of atrial fibrillation using long short-term memory network with {{RR}} interval signals,'' {\em Comput Biol Med}, vol.~102, pp.~327--335, Nov. 2018.

\bibitem{luiMulticlassClassificationMyocardial2018}
H.~W. Lui and K.~L. Chow, ``Multiclass classification of myocardial infarction with convolutional and recurrent neural networks for portable {{ECG}} devices,'' {\em Informatics in Medicine Unlocked}, vol.~13, pp.~26--33, Jan. 2018.

\bibitem{sanninoDeepLearningApproach2018}
G.~Sannino and G.~De~Pietro, ``A deep learning approach for {{ECG-based}} heartbeat classification for arrhythmia detection,'' {\em Future Generation Computer Systems}, vol.~86, pp.~446--455, Sept. 2018.

\bibitem{xiangAutomaticQRSComplex2018}
Y.~Xiang, Z.~Lin, and J.~Meng, ``Automatic {{QRS}} complex detection using two-level convolutional neural network,'' {\em Biomed Eng Online}, vol.~17, p.~13, Jan. 2018.

\bibitem{chenAutomatedArrhythmiaClassification2020}
C.~Chen, Z.~Hua, R.~Zhang, G.~Liu, and W.~Wen, ``Automated arrhythmia classification based on a combination network of {{CNN}} and {{LSTM}},'' {\em Biomedical Signal Processing and Control}, vol.~57, p.~101819, Mar. 2020.

\bibitem{mahmudDeepArrNetEfficientDeep2020}
T.~Mahmud, S.~A. Fattah, and M.~Saquib, ``{{DeepArrNet}}: {{An Efficient Deep CNN Architecture}} for {{Automatic Arrhythmia Detection}} and {{Classification From Denoised ECG Beats}},'' {\em IEEE Access}, vol.~8, pp.~104788--104800, 2020.

\bibitem{strodthoff_ptbxl_benchmark_2020}
N.~Strodthoff, P.~Wagner, T.~Schaeffter, and W.~Samek, ``Deep {Learning} for {ECG} {Analysis}: {Benchmarks} and {Insights} from {PTB}-{XL},'' Apr. 2020.
\newblock arXiv:2004.13701 [cs, stat].

\bibitem{moody_impact_2001}
G.~Moody and R.~Mark, ``The impact of the {MIT}-{BIH} {Arrhythmia} {Database},'' {\em IEEE Engineering in Medicine and Biology Magazine}, vol.~20, pp.~45--50, May 2001.
\newblock Conference Name: IEEE Engineering in Medicine and Biology Magazine.

\bibitem{liu_ICBEB_2018}
F.~Liu, C.~Liu, L.~Zhao, X.~Zhang, X.~Wu, X.~Xu, Y.~Liu, C.~Ma, S.~Wei, Z.~He, J.~Li, and E.~N. Yin~Kwee, ``An {Open} {Access} {Database} for {Evaluating} the {Algorithms} of {Electrocardiogram} {Rhythm} and {Morphology} {Abnormality} {Detection},'' {\em j med imaging hlth inform}, vol.~8, pp.~1368--1373, Sept. 2018.

\bibitem{xiao_classification_review_2023}
Q.~Xiao, K.~Lee, S.~A. Mokhtar, I.~Ismail, A.~L. b.~M. Pauzi, Q.~Zhang, and P.~Y. Lim, ``Deep {Learning}-{Based} {ECG} {Arrhythmia} {Classification}: {A} {Systematic} {Review},'' {\em Applied Sciences}, vol.~13, p.~4964, Jan. 2023.
\newblock Number: 8 Publisher: Multidisciplinary Digital Publishing Institute.

\bibitem{ansari_classification_review_2023}
Y.~Ansari, O.~Mourad, K.~Qaraqe, and E.~Serpedin, ``Deep learning for {ECG} {Arrhythmia} detection and classification: an overview of progress for period 2017–2023,'' {\em Front Physiol}, vol.~14, p.~1246746, Sept. 2023.

\bibitem{zheng_12-lead_CHAPMAN_2020}
J.~Zheng, J.~Zhang, S.~Danioko, H.~Yao, H.~Guo, and C.~Rakovski, ``A 12-lead electrocardiogram database for arrhythmia research covering more than 10,000 patients,'' {\em Sci Data}, vol.~7, p.~48, Feb. 2020.
\newblock Publisher: Nature Publishing Group.

\bibitem{kong_diffwave_2021}
Z.~Kong, W.~Ping, J.~Huang, K.~Zhao, and B.~Catanzaro, ``{DiffWave}: {A} {Versatile} {Diffusion} {Model} for {Audio} {Synthesis},'' Mar. 2021.
\newblock arXiv:2009.09761 [cs, eess, stat].

\bibitem{yuan2024diffusion}
X.~Yuan and Y.~Qiao, ``Diffusion-ts: Interpretable diffusion for general time series generation,'' {\em arXiv preprint arXiv:2403.01742}, 2024.

\bibitem{ronneberger2015unet}
O.~Ronneberger, P.~Fischer, and T.~Brox, ``U-net: Convolutional networks for biomedical image segmentation,'' 2015.

\bibitem{lee2023vector}
D.~Lee, S.~Malacarne, and E.~Aune, ``Vector quantized time series generation with a bidirectional prior model,'' in {\em International Conference on Artificial Intelligence and Statistics}, pp.~7665--7693, PMLR, 2023.

\bibitem{oord2017vqvae}
A.~van~den Oord, O.~Vinyals, and K.~Kavukcuoglu, ``Neural discrete representation learning,'' in {\em Proceedings of the 31st International Conference on Neural Information Processing Systems}, NIPS'17, (Red Hook, NY, USA), p.~6309–6318, Curran Associates Inc., 2017.

\bibitem{wang_time_2016}
Z.~Wang, W.~Yan, and T.~Oates, ``Time {Series} {Classification} from {Scratch} with {Deep} {Neural} {Networks}: {A} {Strong} {Baseline},'' Dec. 2016.
\newblock arXiv:1611.06455 [cs, stat].

\bibitem{optuna_2019}
T.~Akiba, S.~Sano, T.~Yanase, T.~Ohta, and M.~Koyama, ``Optuna: A next-generation hyperparameter optimization framework,'' in {\em Proceedings of the 25th {ACM} {SIGKDD} International Conference on Knowledge Discovery and Data Mining}, 2019.

\bibitem{bynagariGANsTrainedTwo2019}
N.~B. Bynagari, ``{{GANs Trained}} by a {{Two Time-Scale Update Rule Converge}} to a {{Local Nash Equilibrium}},'' {\em Asian j. appl. sci. eng.}, vol.~8, pp.~25--34, Apr. 2019.

\bibitem{liMMDGANDeeper}
C.-L. Li, W.-C. Chang, Y.~Cheng, Y.~Yang, and B.~Poczos, ``{{MMD GAN}}: {{Towards Deeper Understanding}} of {{Moment Matching Network}},''

\bibitem{JMLR:v9:vandermaaten08a}
L.~van~der Maaten and G.~Hinton, ``Visualizing data using t-sne,'' {\em Journal of Machine Learning Research}, vol.~9, no.~86, pp.~2579--2605, 2008.

\bibitem{McInnes2018}
L.~McInnes, J.~Healy, N.~Saul, and L.~Großberger, ``Umap: Uniform manifold approximation and projection,'' {\em Journal of Open Source Software}, vol.~3, no.~29, p.~861, 2018.

\bibitem{JMLR:wang_pacmap_2021}
Y.~Wang, H.~Huang, C.~Rudin, and Y.~Shaposhnik, ``Understanding how dimension reduction tools work: An empirical approach to deciphering t-sne, umap, trimap, and pacmap for data visualization,'' {\em Journal of Machine Learning Research}, vol.~22, no.~201, pp.~1--73, 2021.

\bibitem{liang2024syntheticdatareviews}
W.~Liang, Z.~Izzo, Y.~Zhang, H.~Lepp, H.~Cao, X.~Zhao, L.~Chen, H.~Ye, S.~Liu, Z.~Huang, D.~A. McFarland, and J.~Y. Zou, ``Monitoring ai-modified content at scale: A case study on the impact of chatgpt on ai conference peer reviews,'' 2024.

\end{thebibliography}

\end{document}